\documentclass[letterpaper, 10 pt, conference]{ieeeconf}  

\IEEEoverridecommandlockouts                              

\usepackage{graphicx}
\usepackage{hyperref}
\hypersetup{
    colorlinks=true,
}
\usepackage{amsfonts}
\usepackage{amsmath}
\usepackage{cite}
\usepackage{booktabs}
\usepackage{todonotes}

\title{\LARGE \bf
CVaR-based Flight Energy Risk Assessment for Multirotor UAVs using a Deep Energy Model
}

\author{ Arnav Choudhry$^{*1}$, Brady Moon$^{*2}$, Jay Patrikar$^{*2}$, Constantine Samaras$^{1}$, and Sebastian Scherer$^{2}$
\thanks{*The first three authors contributed equally to this work.}
\thanks{$^{1}$Authors are with the Department of Civil and Environmental Engineering, Carnegie Mellon University, Pittsburgh, PA, USA
        {\tt\small \{arnav, csamaras\}@cmu.edu}}%
\thanks{$^{2}$Authors are with the Robotics Institute, School of Computer Science at Carnegie Mellon University, Pittsburgh, PA, USA
{\tt\small  \{bradym, jpatrika, basti\}@andrew.cmu.edu}}%
}

\begin{document}

\maketitle
\thispagestyle{empty}
\pagestyle{empty}

\begin{abstract}

Energy management is a critical aspect of risk assessment for Uncrewed Aerial Vehicle (UAV) flights, as a depleted battery during a flight brings almost guaranteed vehicle damage and a high risk of human injuries or property damage. Predicting the amount of energy a flight will consume is challenging as routing, weather, obstacles, and other factors affect the overall consumption. We develop a deep energy model for a UAV that uses Temporal Convolutional Networks to capture the time varying features while incorporating static contextual information. Our energy model is trained on a real world dataset and does not require segregating flights into regimes. We illustrate an improvement in power predictions by $29\%$ on test flights when compared to a state-of-the-art analytical method. Using the energy model, we can predict the energy usage for a given trajectory and evaluate the risk of running out of battery during flight. We propose using Conditional Value-at-Risk (CVaR) as a metric for quantifying this risk. We show that CVaR captures the risk associated with worst-case energy consumption on a nominal path by transforming the output distribution of Monte Carlo forward simulations into a risk space. Computing the CVaR on the risk-space distribution provides a metric that can evaluate the overall risk of a flight before take-off. Our energy model and risk evaluation method can improve flight safety and evaluate the coverage area from a proposed takeoff location. \\
The video and codebase are available at: \href{https://youtu.be/PHXGigqilOA}{[Video]\footnote{\url{https://youtu.be/PHXGigqilOA}}} $\mid$ \href{https://git.io/cvar-risk}{[Code]\footnote{\url{https://git.io/cvar-risk}}}\\

\end{abstract}

\section{INTRODUCTION}
Limited battery capacity has been identified as one of the major hurdles in achieving large-scale commercial deployment of Uncrewed Aerial Vehicles (UAVs) \cite{murray2015flying}. While there has been recent growth in improving battery energy density \cite{li2019practical,kerman2017practical}, specific energy \cite{stolaroff2018energy}, and energy management systems \cite{boukoberine2019critical}, limited onboard battery capacity will always pose a significant risk to UAV operations. This risk stems from the difficulty in identifying and quantifying various factors that affect UAV energy consumption. The risk is also a function of the inherent uncertainty in the available estimates of these factors when determining the energy consumption. Traditionally, factors like prevailing wind conditions, air density, payload, UAV design and airspeed \cite{thibbotuwawa2018factors} have been identified as considerably affecting UAV energy consumption. Therefore, a method that quantifies the flight risk based on an estimate of the UAV energy consumption given a certain set of conditions is the focus of this paper. To limit the scope, we specifically focus on multirotor UAV systems but the method is generalizable to any other platform. 

While fixed-wing UAV systems benefit from decades-long research in energy models for fixed-wing crewed aircraft, multirotors can leverage some of the research in rotary-wing aircraft but generally lack such a comprehensive research background by the virtue of the recent development and more complicated physical interactions of UAVs.    
\begin{figure}
    \centering
    \includegraphics[trim={0cm 0cm -.5cm -.3cm},clip,width=.49\textwidth]{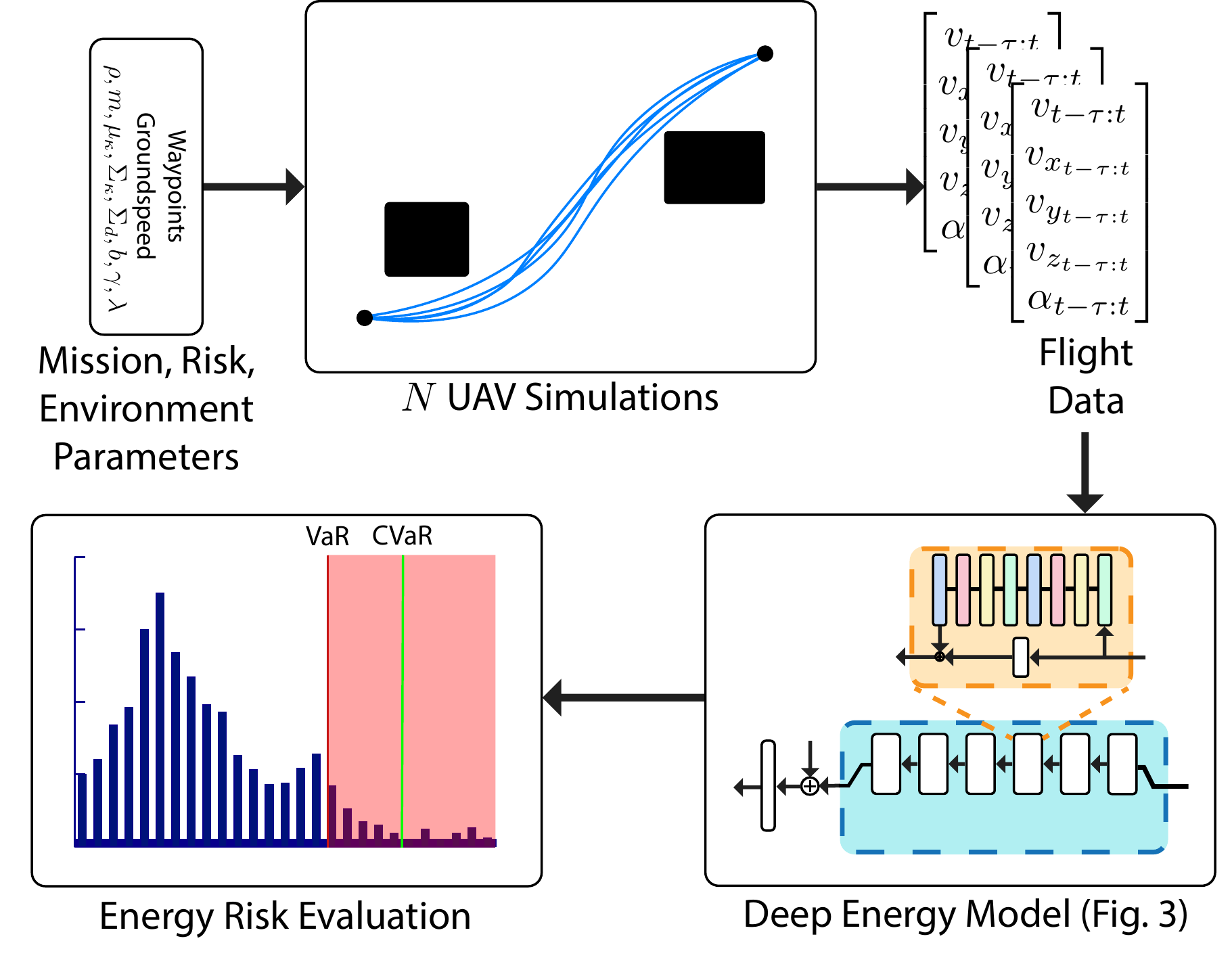} 
    \caption{For a given mission, risk profile, and environmental uncertainty, our algorithm uses a deep energy model in a Monte Carlo forward simulation to generate a energy distribution that informs a CVaR-based risk assessment.}
    \label{fig:full_overview}
\end{figure}

A recent assessment \cite{zhang2020energy} found that while there are multiple attempts at formulating energy model for UAVs, the model estimates differ by a factor of 3-5. The paper attributes this discrepancy to the numerous assumptions in modeling the underlying dynamics of various UAV configurations. The authors of this work argue that the lack of a standardized empirical dataset makes it difficult to compare individual models. Broadly, the prevalent energy models can be grouped into two major categories \cite{prasetia2019mission}. The first are the analytical models that refer to the first-principle based models that rely on established helicopter theories to calculate the energy consumption.
The parameter values for these models are either calculated using historical data \cite{kirschstein2020comparison} or by using a regression-based approach with empirical flight tests \cite{liu2017power}. 
One drawback of using analytical models is that they inherently make simplifying assumptions while formulating the equations. Rotor-Rotor interactions, crosswind effects, and the effect of higher order motions on the propulsion system are often ignored. 
On the other hand, end-to-end approaches completely avoid constructing any explicit models but directly use data-driven methods to regress parameters. The regression is often done by fitting predetermined polynomials to empirical data \cite{abeywickrama2018empirical,maekawa2017power}. Within this paradigm, Prasetia et al. \cite{prasetia2019mission} have presented an approach with multiple polynomial regressions to handle different maneuvers. While the results look promising, the proposed method is complicated and involves generating the movement labels in the dataset to identify various flight maneuvers. The aforementioned methods also do not explicitly take into account external factors like wind or air density which limits their accuracy. A deep learning based model is more expressive than a simple polynomial model. It has been shown recently \cite{bai2018empirical} that compared to other temporal architectures, TCNs are more flexible, stable and less memory intensive.

\par While risk assessment for aerial robots operating in real world is a relatively new research domain, techniques previously developed for performing risk assessments in the domain of applied finance \cite{shapiro2014lectures, rockafellar2000optimization} can be leveraged to perform similar analysis for robot operations. Majumdar and Pavone \cite{majumdar2020should} argue that the traditional risk metrics in robotics like mean-variance may misrepresent the risk associated with a stochastic process. They propose using more informative metrics like Conditional-Value-at-Risk (CVaR) which has been used in some previous works as an optimization constraint to limit the worst-case risk exposure in robotics applications \cite{hakobyan2019risk,sharma2020risk}.

\par In this work, we propose the use of CVaR as a risk metric to quantify the risk associated with worst-case energy consumption. Focusing the analysis on the tail-end of the distribution gives a better idea of the associated worst-case energy use. The energy distribution is obtained by using a Temporal Convolutional Network (TCN) based energy consumption model. The model is trained, validated, and tested on a subset of the energy usage dataset collected using more than 10 hours of real-world flying time \cite{rodrigues2021inflight, Rodrigues}. A Monte-Carlo (MC) forward simulation setup is used to capture the stochasticity of the environment and the UAV dynamics model to capture their effects on the energy consumption. A CVaR value is then generated by mapping the energy distribution to a risk distribution using a user-defined risk profile and battery capacity. Novel case studies using high-fidelity Computational Fluid Dynamics (CFD) wind data using models of real urban areas provide a glimpse on the insights our risk assessment offers. 

\par The major contributions of this work are as follows:
\begin{itemize}
    \item We propose a deep learning based end-to-end model for energy estimation. The model achieves state-of-the-art results for total energy as well as power consumption estimation on a public dataset.
    \item We present a stochastic simulation framework that performs MC runs to provide a distribution on the energy usage given a planned path and environmental conditions.
    \item We provide a CVaR-based risk assessment framework that uses a mapping between the energy distribution to a risk-space using a user-defined risk profile and battery capacity.  
\end{itemize}

\par The paper is organized as follows: Section \ref{sec:energy_model} provides details on the energy model and its results. Section \ref{sec:risk_modeling} provides details on the methods of calculating risk using the energy model and a Monte-Carlo setup. Section \ref{sec:case_studies} gives two case studies to demonstrate applications of the risk method. Section \ref{sec:conclusion} presents the discussions and conclusions. 

\section{Problem Definition}
\label{sec:problem}
Let ${x_t} \in \mathbb{R}^3 \times SO(3)$ represent the state of the UAV in inertial space at time $t$. Let $\xi: [0,t_f] \mapsto \mathbb{R}^3 \times SO(2)$ be a given nominal vehicle trajectory with yaw that the UAV follows using a controller $\pi:(x_t,\xi) \mapsto u_t$ under system dynamics as defined in Equation \ref{eq:dynamics}, where $d$ is the additive noise in system dynamics and $\kappa$ is the environmental context related to features like ambient wind. 

\begin{equation}
\ddot{{x}}_{t} = f({x}_{t},{u}_{t}, \kappa) + {d} \label{eq:dynamics}
\end{equation}

Given battery usable capacity $b$, random variable for energy consumption $e \in E$, where $E \in \mathbb{R}^1_{+} $ is energy-space, our objective is to provide a pre-flight energy-based risk assessment by calculating a suitable risk metric on a empirically derived risk distribution $p(r), r \in R  $ where $R \in  \mathbb{R}^1$ is the risk-space.
\subsection{Approach}
We address the problem of finding the risk distribution and the corresponding metric by first finding an energy distribution $J = \{e|\kappa,{d}\}$ and then transforming it into risk-space using a user-based risk profile $G(e,b): E \mapsto R$. To do this, we first construct a power consumption model $ Q({\eta}_{t-\tau:t})$, where ${\eta}_{t-\tau:t}$ is a set of derived features from state history ${x}_{t-\tau:t}$ and $\kappa$. We then use $N$ MC forward simulations to find the energy distribution of a UAV while flying the trajectory $\xi$, under a pre-defined distribution of environmental noise $\kappa \sim N(\mu_{\kappa},\Sigma_{\kappa})$ (epistemic uncertainty) and a distribution of system dynamics noise ${d} \sim N(0,\Sigma_{d}) $ (aleatoric uncertainty), using Eq. \ref{eq:energy}.
\begin{equation}
e = \int_0^{t_f} Q({\eta}_{t-\tau:t}) dt
\label{eq:energy}
\end{equation}
The MC generates a energy histogram with M bins which is then transformed from energy-space to a risk-space using the risk-profile $G(e,b)$ to approximate $\hat{p}_N(r)$. The risk-probability for $r$ in a given bin $R_l$ is given by Eq. \ref{eq:bin}.
\begin{equation}
    \hat{p}_N(r) =  \frac{M}{N}\sum_{i=1}^N I(G(e,b) \in R_l) 
    \label{eq:bin}
\end{equation}

The risk-distribution can now be analysed using metrics like Conditional Value-at-Risk (CVaR) which the authors argue is a relevant metric in that it effectively captures the risk posed by the tail-end distribution. The CVaR value can then be used as part of a comprehensive evaluation that combines other risk factors such as risk to property and people on the ground to aid in a go/no-go pre-flight decision.
An overview of our approach can be seen in Figure \ref{fig:full_overview}.

\section{Energy Modeling}
\label{sec:energy_model}
This section details the construction of the power consumption model $Q$. In order to estimate the power consumed by the UAV, we introduce an end-to-end model that takes time varying features and invariant features, and estimates the power consumption of the UAV at each time step. This is done by framing the problem as a multivariate time-series regression problem.
\subsection{Energy Data Description}

\begin{figure}
    \centering
    \includegraphics[trim={0cm 0cm -.9cm -.5cm},clip,width=0.48\textwidth]{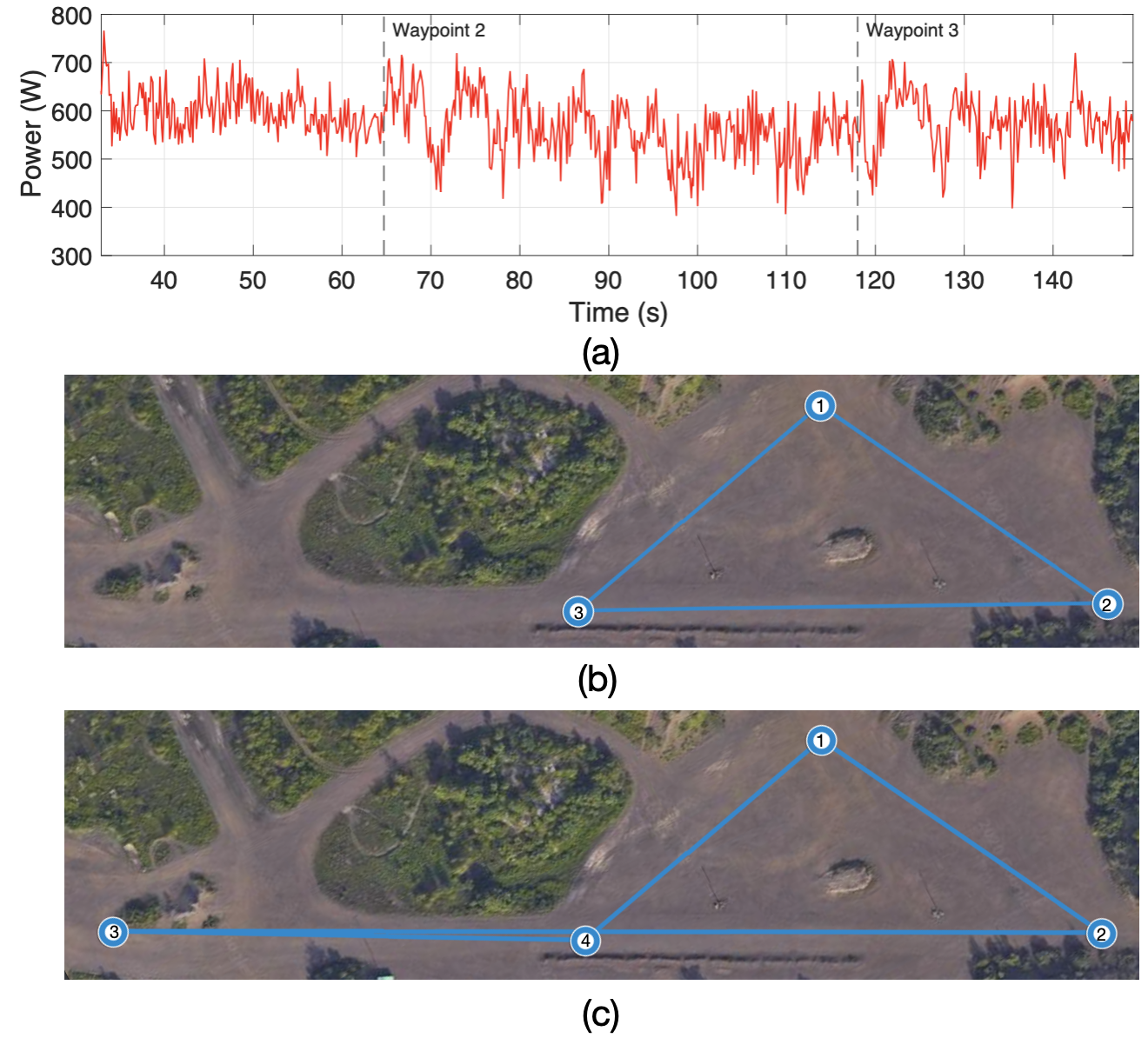}
    \caption{Figure shows the dataset used for training the energy model. The top image (a) shows power output for the GPS route (b) that is used for the majority of the dataset. The bottom image (c) shows the route used to create unique variations for the testing set.}
    \label{fig:flight_paths}
    \vspace{-3.5mm}
\end{figure}

We use a publicly available dataset \cite{rodrigues2021inflight, Rodrigues} which contains power and state measurements for 209 flights on a DJI Matrice M100 drone. 195 of these flights were performed while following a fixed triangular trajectory (see Figure \ref{fig:flight_paths}b) with varying altitudes, speeds, and payloads, while the remainder of 10 flights (referred to as random flights) followed a different flight path (see Figure \ref{fig:flight_paths}c) for unique testing data. We split the flights as 60:20:20 for the train, validation and test sets of flights, while making sure that the 10 random path flights were only in the test set. We used a total of 114 training flights (4 hours and 4 minutes flight time), 38 validation flights (1 hour and 24 minutes flight time) and 37 test flights (1 hour and 23 minutes flight time).
\subsection{Energy Model Architecture}

\begin{figure}
    \centering
    \includegraphics[trim={0cm 2.5cm 0.0cm -.5cm},clip,width=.48\textwidth]{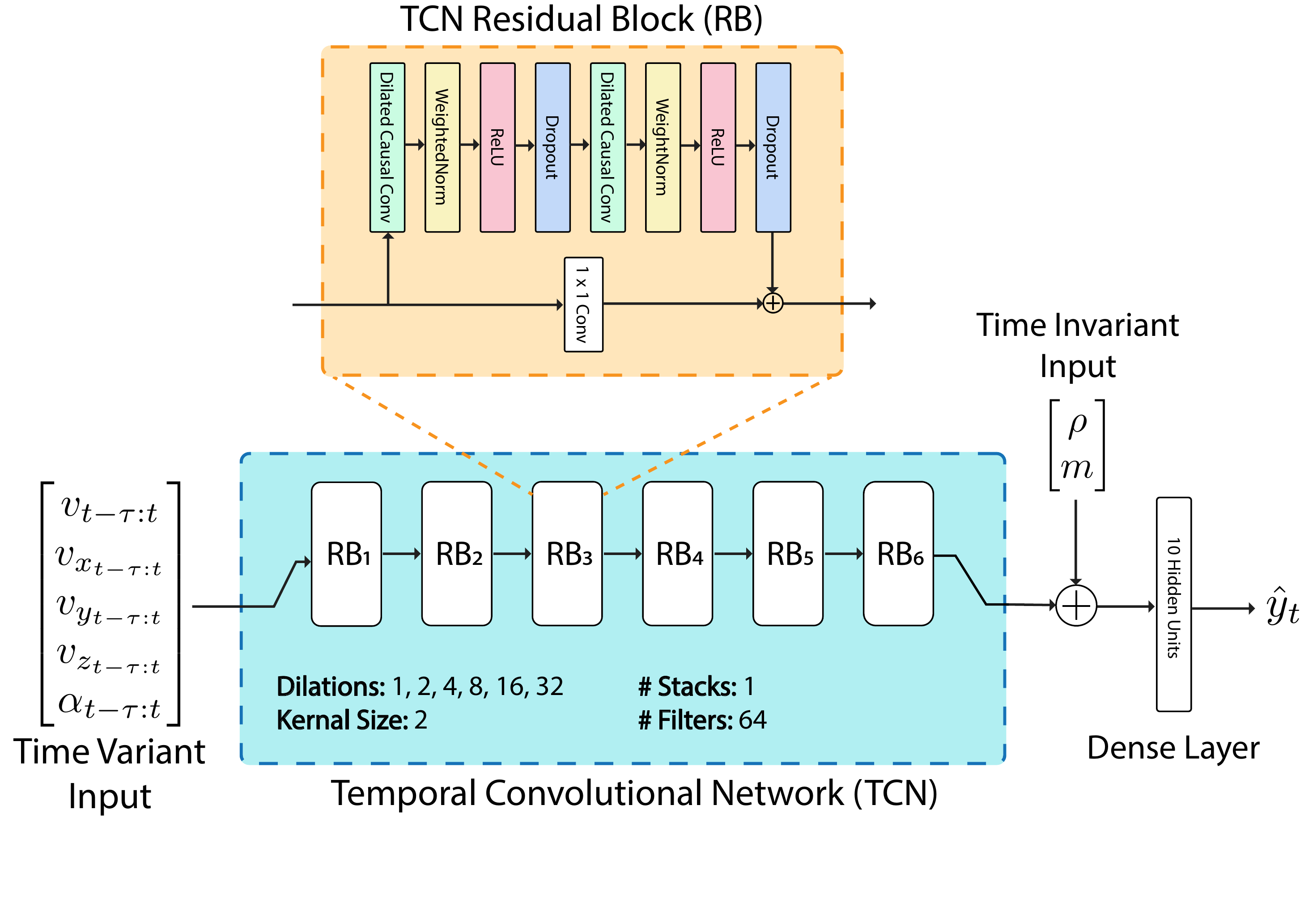} 
     \caption{Overview of the deep energy model. The time varying features are passed into the TCN, whereas the time invariant features are concatenated to the output of the TCN and passed through a dense layer. The hyperparameter values correspond to b-TCN.}
    \label{fig:deep_net}
\end{figure}


Figure \ref{fig:deep_net} provides an overview of the deep energy model. We construct and split $\eta_t$  as the time varying features $\eta^v_t$ and time invariant features $ \eta^i $.  
The time varying features $ \eta^v_t = \{ v_t , v_{x_t}, v_{y_t}, v_{z_t}, \alpha_t \} $ are airspeed $v_t$, its components in body-frame $v_{x_t}$ and $v_{y_t}$, vertical speed $v_{z_t}$, and the angle of attack $\alpha_t$ \footnote{The angle of attack is approximated to pitch angle as the dataset does not include explicit angle of attack information.}. The time-invariant features $ \eta^i = \{ \rho, m \}$ are the density of air $\rho$, and the mass of the payload $m$.
A vector $\Phi_t$ containing temporal embedding of features is obtained from the temporal encoder.

The contextual information is then added to $\Phi_t$ by concatenating time-invariant features to create an input $\Phi'_t = \Phi_t \oplus \eta^i $, containing both temporal as well as contextual information.  $\Phi'_t$ is then fed into a dense layer to estimate the power consumption of the UAV at time $t$. There are two reasons we separated the time-varying and fixed inputs. The first is we noticed that separating the inputs improves the convergence rates, and second is that it makes the model modular.

In order to encode the time-varying information, we chose a TCN-based \cite{oord2016pixel} architecture. In the interest of comparative analysis, we also tested LSTMs \cite{hochreiter1997long} for the temporal modeling. 
For the TCNs, we specifically used the implementation as described in \cite{bai2018empirical}. TCNs have been shown to perform at least at or better than LSTMs in a variety of temporal modeling tasks with added advantages including lower memory requirements, flexible receptive field sizes, and stable gradients.

\subsection{Model Experiments}


We perform a hyperparameter search for both TCNs and LSTMs. For TCNs, we vary the number of filters in each causal convolution unit between $\{16, 32, 64, 128\}$, the kernel size between $\{2, 3, 4\}$, the number of residual block stacks between $\{1, 2\}$, and the number of layers between $\{3, 4, 5, 6\}$. We use the Adam optimizer with a learning rate of 0.002 and make use of exponential dilation \cite{bai2018empirical}. We selected the top two models LSTM and TCN models based on their performance on the validation set and evaluate them on the test set. The results of which are in Table \ref{tab:comparison} and are further discussed in Section \ref{deep_results}.

\subsection{Metrics}
We use the mean absolute percentage error (MAPE) to measure the performance of the models on the power estimation task. The MAPE for a flight $i$, is defined in Eq. \ref{eq:mape}, where $t=[0,t_f]$, and $y_t^{(i)}$ and $\hat{y_t}^{(i)}$ are the true and predicted value of power for flight $i$ at time step $t$, respectively. 
\begin{equation}
\label{eq:mape}
    \text{MAPE}_i = \frac{1}{t_f^{(i)}} \sum_{t=0}^{t_f^{(i)}} \left|\frac{y_t^{(i)} - \hat{y}_t^{(i)}}{y_t^{(i)}}\right| \times 100\%
\end{equation}

 For measuring the model performance on the energy estimation task, we first divide the flight into sections, $s \in \{1, \ldots, S^{(i)}\}$, based on the yaw angle. We then calculate the energy consumption in each section, $e_s^{(i)}$, and use these to calculate the adjusted relative error for flight $i$, $\text{RE}_i$ in Eq.  \ref{eq:RE}, where $\hat{e}_s^{(i)}$ is the predicted energy for each section. 
\begin{equation}
\label{eq:RE}
\text{RE}_i = \frac{1}{S^{(i)}} \sum_{s=1}^{S^{(i)}} \left|\frac{e_s^{(i)} - \hat{e}_s^{(i)}}{e_s^{(i)}}\right| \times 100\%
\end{equation}
This metric forces the model to accurately predict total energy consumed in all portions of the flight and not have errors from different parts of a flight cancel out.

\subsection{Energy Model Results} \label{deep_results}
\begin{figure}
    \centering
    \includegraphics[trim={0cm 0cm 1.5cm -1.5cm},clip,width=0.49\textwidth]{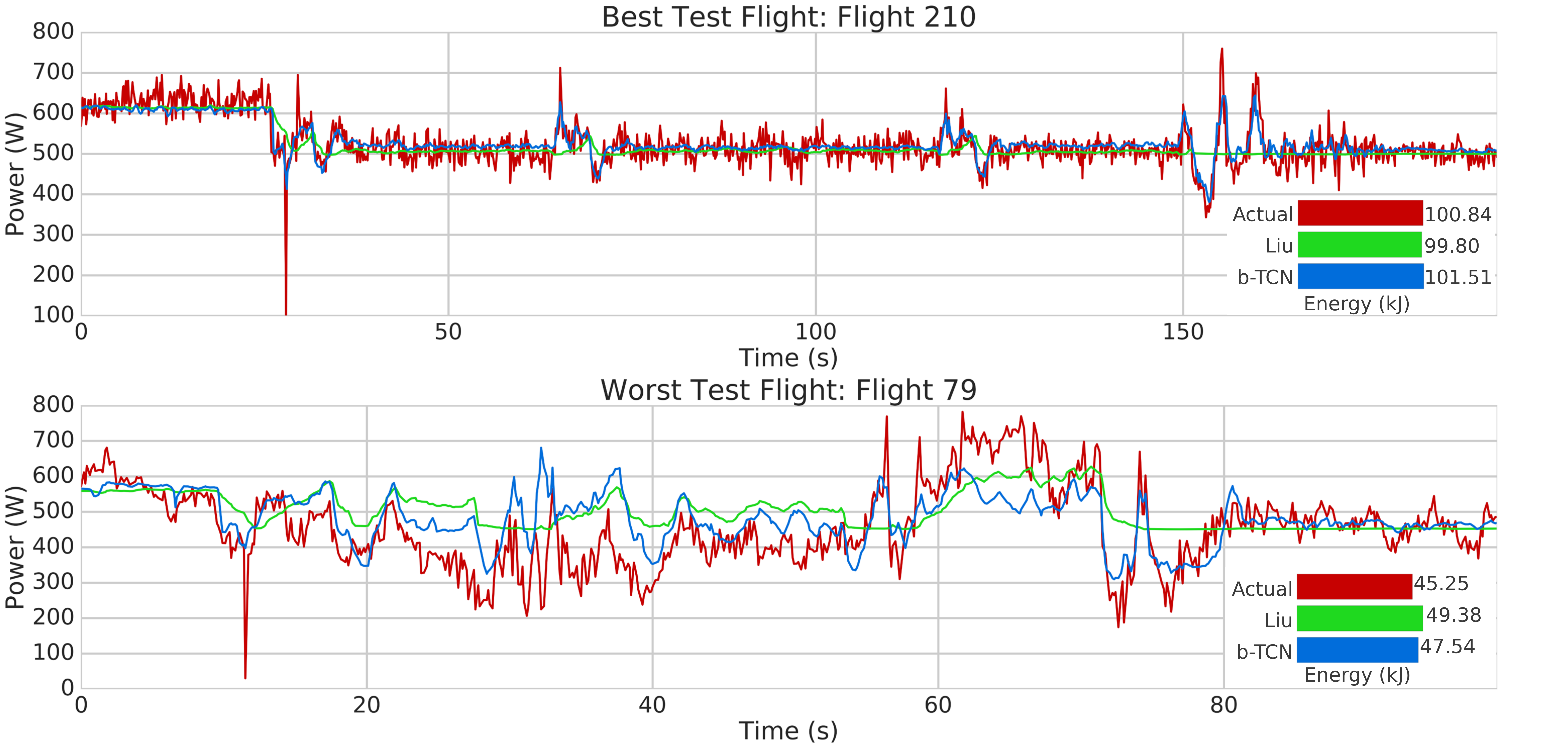} 
    \caption{Test Flights with the best and worst MAPE for b-TCN. The bar graphs in the bottom right corners represent the total energy estimate over the entire flight.}
    \label{fig:comparison}
\end{figure}
\begin{table}
    \centering
    \caption{Energy Model Results}
    \vspace{-2.5mm}
    \begin{tabular}{lccccr}
    \toprule
         \textbf{Model} & \multicolumn{2}{c}{\textbf{Random Flights}} & \multicolumn{2}{c}{\textbf{Test Flights}} & \textbf{\# params} \\
     \cmidrule{2-5}
          & MAPE(\%) & RE(\%) & MAPE(\%) & RE(\%) & \\
    \midrule
    Liu \cite{liu2017power} &  14.05 & 9.16 &  12.70 & 7.41  &   5 \\
    b-LSTM &  13.58 & 10.35 &  11.47 & 7.57  &  84,649 \\
    s-LSTM &  15.35 & 11.65 &  12.36 & 8.24  &  5,833 \\
    b-TCN &   10.38 & \textbf{7.38} &  \textbf{9.06} & \textbf{5.12}  &   76,073 \\
    s-TCN &   \textbf{10.36} & 7.56 &  9.13 & 5.52 &  5,225 \\
    \bottomrule
    \end{tabular}
    \label{tab:comparison}
    \vspace{-2.5mm}
\end{table}

Based on their performance on the validation set, we selected the top two LSTM and TCN models. The \emph{b-TCN} model is the bigger TCN model with 64 filters, kernel size 2, 1 residual block stack, and 5 layers. The \emph{s-TCN} model is the smaller TCN model with 16 filters, kernel size 2, 1 residual block stack, and 5 layers. Both \emph{b-LSTM} and \emph{s-LSTM}, had 3 hidden layers with 0 dropout and were trained using RMSprop optimizer. b-LSTM had 64 hidden units, whereas s-LSTM had 16 hidden units.

From Table \ref{tab:comparison}, we see that the TCNs consistently outperform the LSTMs on the power as well as energy estimation tasks. Additionally, note that even though the s-TCN is about 93\% smaller than the b-TCN, the performance of both is similar. This shows that the TCN architecture lends itself better to model compression and hence might be better for edge applications by using less memory and faster inference. However, it should be noted that the s-TCN takes considerably more number of epochs to train. Figure \ref{fig:comparison} shows time-series plots of the best and worst flights for the b-TCN. We can see that the model is able to follow the peaks and troughs of power consumption. 
\section{Risk Modeling}
\label{sec:risk_modeling}

\subsection{Wind Modeling}
\label{sec:wind}

An important factor in predicting energy consumption for a UAV is accounting for the effects of the wind field. In order to accurately simulate the energy usage of a given trajectory, an accurate wind model must be included in the environmental context $\kappa$ for the system dynamics of Equation \ref{eq:dynamics}. For this work, wind is broken down into two components: constant wind and turbulence.

The constant wind is the averaged wind field which is temporally static but spatially varying for a given environment. Given an accurate model of the environment, the current wind field can be found using the wind inlet angle $\Theta$ and magnitude $U$ by using CFD. 
CFD has been shown to be an effective method in calculating the low altitude urban wind flow patterns for UAV flight \cite{orr2005framework, ware2016analysis, galway2011modeling, pat2020}. However, wind fields are not temporally constant because the inlet conditions are not steady and therefore need to be defined as a distribution. The mean and standard deviations of the inlet angle and magnitude are given as $\mu_{\kappa} = \{\mu_\Theta,\mu_U\}$ and $\Sigma_{\kappa} = diag(\sigma_\Theta,\sigma_U)$ respectively and can come from historical data spanning over years or even an hour depending on the application. 

In addition to the constant wind, turbulence is added using the Dryden turbulence model \cite{mil}. 
Previous works \cite{abichandani2020wind,wasland2009, allison2018, Beard2012SmallUA} have shown that this model can be adapted for simulation of wind turbulence for small UAV.

\subsection{Monte Carlo Simulations}
\label{sec:mc}

Given $\mu_{\kappa}$, $\Sigma_{\kappa}$, and $\Sigma_{d}$, we run $N$ MC simulations to create a set of possible vehicle trajectories given the distribution of conditions. 
The simulated vehicle follows the trajectory $\xi$ with the controller $\mathbf{\pi}$ and the dynamics defined in the previous section.
\begin{figure*}
    \centering
    \includegraphics[trim={0cm 0cm 0.0cm -.4cm},clip,width=0.9\textwidth]{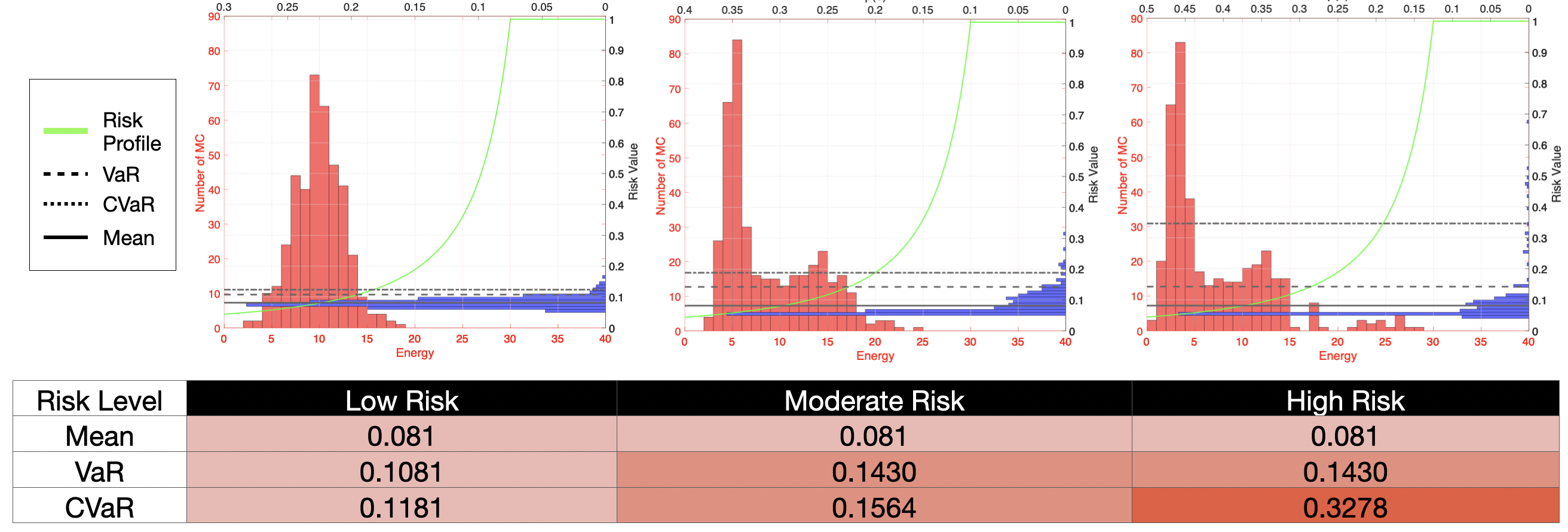}
    \caption{Illustrative example to show how CVaR is a more relevant risk metric than mean and VaR as it captures the tail-end distribution better. }
    \label{fig:cvar_risk}
    \vspace{-2.5mm}
\end{figure*}


\subsection{Risk Modeling}
\label{sec:risk}

As defined in Section \ref{sec:problem}, $G(e,b)$ represents the user-defined operation risk profile. In this work, as a test case, $G(e,b)$ can be chosen as in Eq. \ref{eq:risk_profile}.
\begin{equation}
    G(e,b) = \exp\Bigg(\frac{\gamma}{(\max\{b-e,\lambda\})}\Bigg)-1
    \label{eq:risk_profile}
\end{equation}
where $\gamma, \lambda > 0$ are constants. Such a choice of $G(e,b)$ takes into account the fact that as the energy consumption nears the battery capacity, the risk increases exponentially until it reaches a limit value. 
\par The MC simulations defined in the previous section can be used to generate the distribution for $e$ which in turn can be used to generate the risk probability density function $p(r)$ by normalizing the output of $G(e,b)$. From here on we follow the formulation proposed by Rockafellar et al. \cite{rockafellar2000optimization} to define Conditional Value-at-Risk (CVaR).


The Conditional Value-at-Risk at level $\nu$ or CVaR$_{\nu}$ gives the expected risk of the worst $(1-\nu)$ quantile of the distribution $G(e,b)$. Intuitively CVaR$_{\nu}$ gives us an idea on the behaviour of the worst $(1-\nu)$ quantile cases by providing the expected risk in the cases that are ``beyond'' Value-at-Risk at level $\nu$ (VaR$_{\nu}$) risk.
\begin{equation}
    \text{CVaR}_{\nu} = \frac{1}{1-\nu}\int_{G(e,b) \geq \text{VaR}_{\nu}} G(e,b)p(r)dr
    \label{eq:cvar}
\end{equation}


While the mathematical superiority of CVaR$_{\nu}$ as a \textit{coherent risk metric} \cite{majumdar2020should} is useful in formulating convex CVaR$_{\nu}$ optimization problems \cite{rockafellar2000optimization}, it also helps to intuitively see how VaR$_{\nu}$ and CVaR$_{\nu}$ help provide information about a particular risk distribution profile. Figure \ref{fig:cvar_risk} shows a deliberately constructed example that highlights how the different metrics---mean, VaR$_{\nu}$, and CVaR$_{\nu}$---capture the risk profile. The three figures show three different energy MC outputs in red. These outputs are then transformed into risk space by using equation \ref{eq:risk_profile} with coefficients $\gamma = 6.93$, $\lambda = 10$, $b = 40$ J. This in turn gives us $p(r)$ or the risk $p.d.f$ in blue. VaR$_{\nu}$ and CVaR$_{\nu}$ values are calculated at $\nu = 0.95$. As can be seen, while intuitively it is clear that moving to the right of the three examples, the MC energy output relates to a higher risk flight because of a longer tail, both the mean and VaR$_{\nu}$ are not able to capture this behaviour. 

\section{Case Studies}
\label{sec:case_studies}

In order to illustrate potential applications of the presented work, two case studies will be given. The first case study shows how the energy risk for a proposed flight trajectory would be evaluated and could be used before the fly or no-fly command is given. The second case study is an evaluation of all potential flight paths from a given takeoff location to evaluate the coverage area for a proposed takeoff location. 


\subsection{Case 1: }
Before a UAV takes off to carry out a nominal path, it would be prudent to evaluate the risk of depleting the battery during flight. Using our proposed method of evaluating risk, the pilot in command could calculate the distribution of necessary energy to carry out the flight path and its CVaR$_{\nu}$. The environmental conditions for the MC simulations would model the conditions during the proposed flight window.

To demonstrate this case, real environmental data is used from a test site with around ten buildings ranging in heights of 5 to 35 meters. This area is similar to a typical urban environment with non-uniform wind fields. The area is around $300 \times 250$ meters. The wind field model was calculated using CFD, and the wind inlet conditions were captured using a weather station on the tallest building in the area. The window of time for this test was February 19, 2020 from 4:20-4:22 PM. The inlet conditions were $\mu_{\kappa} = \{-2.53^\circ,3.14 m/s\}$, $\Sigma_{\kappa}  = diag(28.47^\circ, 1.55 m/s) $

\begin{figure}
    \centering
      \includegraphics[trim={0cm 0cm 0cm -.4cm},clip,width=.99\linewidth]{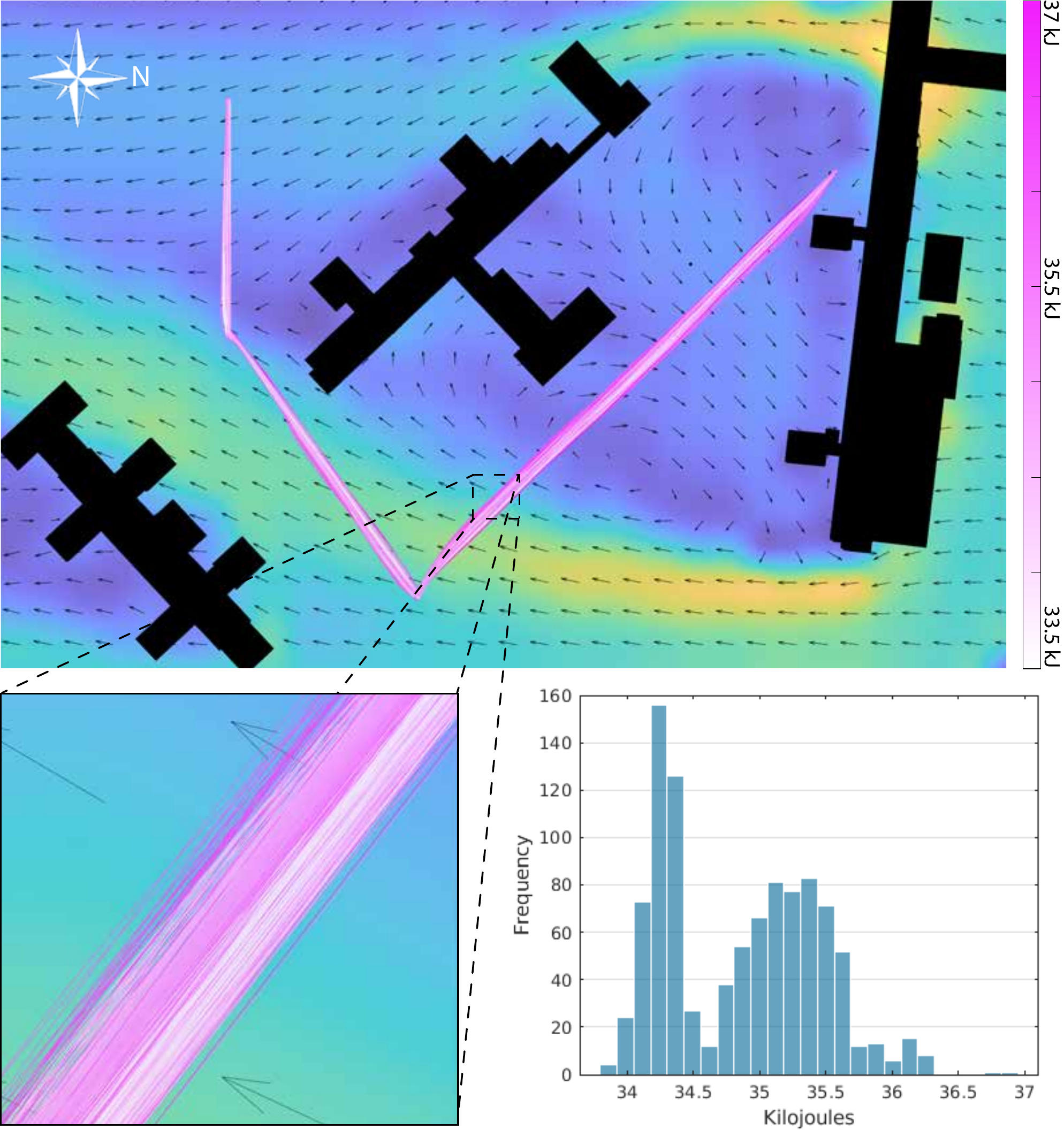}
    \caption{Monte Carlo simulations for Case 1. Each flight is represented with a pink line and is shaded according to the total energy consumption. The distribution of the total energy consumption for the flights is shown in the histogram.}
    \label{fig:study1}
\end{figure}

Figure \ref{fig:study1} shows the $1000$ MC simulations and the predicted amount of energy for the trajectory. The mean energy consumption was $34.86$ kJ. With $b=369.36$ kJ, $\lambda=92.34$, $\gamma=64$, and $\nu=.95$, the CVaR$_{\nu}$ value is $0.2117$. This value reflects how risky this proposed flight path is and would be compared against the allowable risk determined by the operator or company.


\subsection{Case 2: }
In many applications of a UAV, the vehicle will take off and return to the same location every time. In these cases it could be useful to evaluate the coverage area for that particular location. This knowledge could influence the placement of launching location and optimize the number of base stations necessary to ensure full coverage of a region.

The testing environment for this case is a $3300 \times 2130$ meter area of a major metropolitan city. This city has a higher density of buildings compared to Case 1. The wind field model for this city came from a high resolution CFD simulation. One starting location is chosen at a point close to the center of the map, and over 1000 goal locations are randomly sampled from the space within a given radius. A BIT* planner \cite{Gammell_2015} finds a path connecting the start and goal states. For this example we used yearly wind statistics as the inlet conditions in order to evaluate the coverage area over a long period of time. Flights for each path are then simulated and evaluated using our risk assessment method.

\begin{figure}
    \centering
      \includegraphics[trim={.47cm 2.5cm 0.75cm 3.4cm},clip,width=.99\columnwidth]{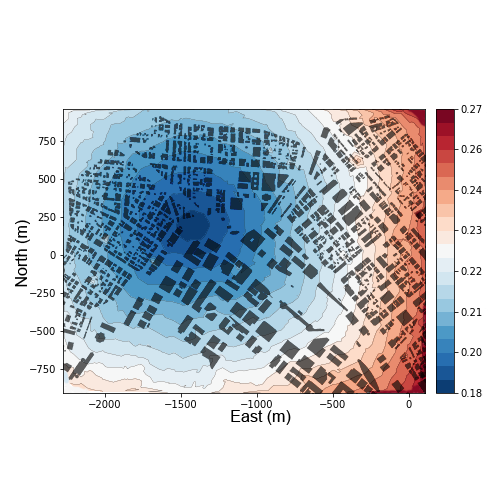}
    \caption{Contour plot of the CVaR$_\eta$ values for Case 2 with the city map overlay.}
    \label{fig:study2}
\end{figure}

Figure \ref{fig:study2} shows a contour plot of the CVaR$_\eta$ to reach each location in the region. Natural corridors lead to lower energy consumption and thus lower risk values. In contrast, regions of the map with dense building placements leads to longer paths, higher energy consumption, and higher risk.
With a given risk value limit, the coverage area can be calculated for the given environment conditions.
\begin{figure}
    \centering
      \includegraphics[trim={.47cm 2.5cm 0.55cm 3.4cm},clip,width=.99\columnwidth]{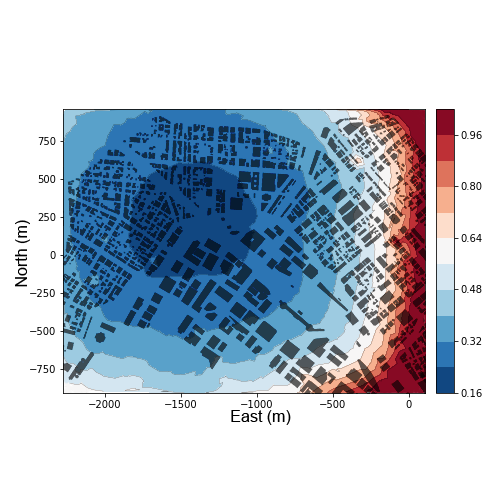}
    \caption{Contour plot of the CVaR$_\nu$ for Case 2 when the usable battery capacity is low. The risk values increase quickly as the UAV travels away from its takeoff location.}
    \label{fig:study2bat}
\end{figure}

With an adjusted risk profile to reflect a lower usable battery capacity, the contour plot of the CVaR$_\nu$ values changes dramatically, as can be seen in Figure \ref{fig:study2bat}. The risk values steadily increase when traversing farther away from the takeoff location and even approaches a risk value of 1.


\section{Conclusions}
\label{sec:conclusion}
This work presents a novel, open-source, state-of-the-art, deep-learning based energy model trained on a publicly available dataset to quantify the energy risk for multirotor UAVs. MC forward simulations are used to find a energy distribution which is transformed into a risk-space. CVaR is used to conduct a risk assessment on the risk-space distribution. Case studies using real world data highlight how the proposed framework can be applied. 

While the proposed deep energy model is generalizable across various multirotor configurations, we only show results using a quadrotor model. The results for the energy model are also only tested in typical delivery-style missions and do not account for aggressive, evasive, or acrobatic maneuvers. Even with these limitations, we demonstrate that CVaR-based risk assessment can be used for energy consumption analysis of multirotor systems.    








\section*{ACKNOWLEDGMENT}
This work is supported by the U.S. Department of Energy (Grant DE-EE0008463). This material is based upon work supported by the National Science Foundation Graduate Research Fellowship under Grant No. DGE1745016.

The authors would like to thank Bastian Wagner for his help with the implementation of the analytical model and Cherie Ho for her help with the MC simulations.

\bibliography{ref}

\end{document}